\pdfoutput=1

\documentclass[11pt]{article}

\usepackage[final]{acl}

\usepackage{latexsym}

\usepackage[T1]{fontenc}

\usepackage[utf8]{inputenc}
\usepackage{url}
\usepackage{hyperref}

\usepackage{microtype}
\usepackage{booktabs} 
\usepackage{times} 
\usepackage{xcolor} 
\usepackage{multirow}
\usepackage{scalerel} 
\usepackage{graphicx}
\graphicspath{    
    {imgs} 
}

\usepackage{tabularx}
\usepackage{soul}
\definecolor{lightblue}{rgb}{.88,.98,1}
\definecolor{lavendermist}{rgb}{0.9, 0.9, 0.98}
\definecolor{lemonchiffon}{rgb}{1.0, 0.98, 0.8}
\sethlcolor{lemonchiffon} 

\usepackage{wrapfig,lipsum}
\newcommand\mC[1]{\multicolumn{1}{c}}

\usepackage{titlesec}
\titleformat{\section}{\normalfont\large\bfseries\center}{\thesection.}{1em}{}
\titleformat{\subsection}{\normalfont\fontsize{11}{12}\bfseries\raggedright}{\thesubsection.}{1em}{}
\titleformat{\subsubsection}{\normalfont\normalsize\bfseries\raggedright}{\thesubsubsection.}{1em}{}
\renewcommand\thesection{\arabic{section}}
\renewcommand\thesubsection{\thesection.\arabic{subsection}}
\renewcommand\thesubsubsection{\thesubsection.\arabic{subsubsection}}

\usepackage{epstopdf}
\usepackage[utf8]{inputenc}

\usepackage{hyperref}
\usepackage{xstring}

\usepackage{array}
\newcolumntype{P}[1]{>{\centering\arraybackslash}p{#1}}

\usepackage{color}
\usepackage{csquotes}

\usepackage{amsmath}
\usepackage{amssymb}
\usepackage{amsfonts}
\usepackage{pifont}
\newcommand{\cmark}{\ding{51}}
\newcommand{\xmark}{\ding{55}}
\newcommand{\BM}{{\cal BM\text{25}}}

\title{Divide \& Conquer for Entailment-aware Multi-hop Evidence Retrieval}

\author{Fan Luo \\
  University of Arizona \\
  Tucson, AZ, USA \\ 
  \texttt{fanluo@email.arizona.edu} \\\And
  Mihai Surdeanu \\
  University of Arizona \\
  Tucson, AZ, USA \\ 
  \texttt{msurdeanu@email.arizona.edu} \\}

\begin{document}
\maketitle
\begin{abstract}

Lexical and semantic matches are commonly used as relevance measurements for information retrieval. Together they estimate the semantic equivalence between the query and the candidates. However, semantic equivalence is not the only relevance signal that needs to be considered when retrieving evidences for multi-hop questions. In this work, we demonstrate that textual entailment relation is another important relevance dimension that should be considered. To retrieve evidences that are either semantically equivalent to or entailed by the question simultaneously, we divide the task of evidence retrieval for multi-hop question answering (QA) into two sub-tasks, i.e., semantic textual similarity and inference similarity retrieval. We propose two ensemble models, EAR and EARnest, which tackle each of the sub-tasks separately and then jointly re-rank sentences with the consideration of the diverse relevance signals. Experimental results on HotpotQA verify that our models not only significantly outperform all the single retrieval models it is based on, but is also more effective than two intuitive ensemble baseline models. 

\end{abstract}

\section{Introduction}

Widely adopted QA approaches use a two-stage pipeline, i.e., a retriever module followed by a reader module \citep{chen2017reading}. The retriever is responsible for collecting relevant  evidences, then the reader module combines the relevant information from the retriever module to infer the answer. Evidence retrieval is a ranking task, which usually maps questions and candidates to vectors and scores the semantic relationship between them with respect to relevance. The goal is to rank the most relevant pieces of evidence (supporting facts) at the top of the list among all the candidates.

According to formal semantic notions, the semantic relationship between two text fragments 
includes semantic equivalence, referential equality, and textual entailment. While referential equality can be mostly solved by coreference resolution and entity linking, semantic similarity and textual entailment require deep semantic understanding between question and context \citep{lin2021pretrained}. Textual entailment is a framework that captures semantic inference. Textual Entailment (TE) of two text fragments can be defined as the task of deciding whether the meaning of one text fragment can be inferred from another text fragment. That is, a premise T entails a hypothesis H if, typically, a human reading T would infer that H is most likely true. For example, 
      {\it ``T: Jack sold the house to Peter.
      H: Peter owns the house.}
Here H can be inferred from T, so T entails H. 
While semantic equivalence capture ``Do these two texts mean the same thing?'', textual entailment is a framework that captures semantic inference, deciding whether the meaning of one text can be inferred from another.

\begin{figure}[h!]
\centering
\input{imgs/example}
\caption{\label{tab:Examples}
An example from the HotpotQA dataset showing the two different dimensions of relevance between the question and its supporting evidences.
}
\label{fig:bigexamples}
\end{figure}

It is common to utilize an inference model as a reader to infer the correct answer from a subset of retrieved context. However, most works of the evidence ranking only measure the semantic textual similarity between the question and candidate corpus to determine the relevance
, and ignore the inference signals.
This works for single-hop questions, in which relevant information usually share the same entity mentions with the question. However, it is not sufficient for multi-hop questions.  
The relevance between multi-hop question and its evidence(s) is beyond the lexical or semantic similarity that is targeted by most retrievers, especially for secondary hops.  
As the example in Figure~\ref{fig:bigexamples} shows, relevance is multi-dimensional. The first hop evidence {\it ``James Henry Miller (25 January 1915 – 22 October 1989), better known by James Henry Miller stage name Ewan MacColl, was an English folk singer, songwriter, communist, labour activist, actor, poet, playwright and record producer"} has lexical overlap with the question and thus get high scores from token overlap match and semantic similarity comparison. 
However, the subsequent hops in the paragraph {\it `Peggy Seeger'} failed to be retrieved by both lexical match and semantic similarity comparison, as they would not be considered as ``mention or mean the same thing'' as the question. Instead, they have an entailment relation with the question. {\it ``Margaret `Pegg' Seeger (born June 17, 1935) is an American folksinger.''} entails a nationality relationship, and  similarly {\it `James Henry Miller’s wife'} entails a marriage relationship. 
Thus, we argue that semantic similarity is not the only relevance signal that needs to be considered when ranking evidences for multi-hop questions.

To consider the fine-grained aspects of relevance for the multi-hop QA evidence retrieval task, we divide the task into separate retrieval subtasks, where each aims to retrieve a subset of sentences that score highly on one of the relevance dimension (i.e., semantic equivalence or textual entailment) respectively, and then combine them to output the final ranking with an ensemble model.

Our contributions are:
1. We call attention that fine-grained aspects of a model’s capability in performing types of relevance signals should be considered. Especially, textual entailment should be explicitly taken into consideration when measure relevance for complex question answering evidence retrieval, so that the retriever covers a more accurate relevant context that not only lexical match based but also require inference to identify. 
2. we propose two ensemble models that combine diverse relevance signals captured by different base models. Our experimental results demonstrate that not only are the individual base retrieval model necessary in evidence retrieval but cooperate advantageously to produce a better ranking for multi-hop QA evidence retrieval when used together .
3. We empirically show the effectiveness of the proposed ensemble retrieval models by evaluating on multi-hop HotpotQA dataset and show they not only outperform all the base models, and also several ensemble baselines.

\section{Related Work}

\subsection{Text Retrieval}

\textbf{Traditional retrieval models} such as TF-IDF and BM25 \citep{trotman2014improvements} use sparse bag-of-words representations to collect lexical matching signals (e.g., term frequency).
Such sparse retrieval models are mostly limited to exact matches.
\textbf{Dense Retrieval models} 
move away from sparse signals to dense representations, which
help address the vocabulary mismatch problem. These models can be categorized into two types according to their model architecture, \textit{representation-based} \citep{huang2013learning, shen2014learning} and \textit{interaction-based} models \citep{pang2016study,lu2013deep, guo2016deep,mitra2017learning}. 
\textbf{Hybrid methods} \citep{lin2021pyserini, gao2020complementing,karpukhin2020dense, shan2020bison} aggregate the dense retrieval with sparse retrieval methods to better model relevance. 
The entailment-aware retrieval models we propose are also hybrid methods that combine sparse and dense retrieval methods, but our method is unsupervised.
Further, we combine a sparse model with multiple dense models to consider diverse relevance signals, i.e., textual entailment in addition to semantic equivalence.

\subsection{Multi-hop Evidence Retrieval}
Research works on multi-hop evidence retrieval can be broadly categorized into two directions:
\textbf{(1) Question decomposition:} \citep{min2019multihop, jiang2019self, fu2021decomposing, perez2020unsupervised, talmor2018web} 
decompose multi-hop questions into multiple single-hop sub-questions.
The question decomposition works that relies on a decomposer model trained via supervised labels to be able to decompose the multi-hop questions into single-hop questions accurately. In contrast, our entailment-aware retrieval models tackle this task in the opposite direction. That is, instead of decompose the question, we assemble candidate sentences pairs that carrying different relevance signals to match against the question. 
\textbf{(2) Iterative evidence retrieval:} \citet{feldman2019multihop} proposed a method to iteratively retrieve supporting paragraphs, using the paragraphs retrieved in previous iteration to reformulate the search vector. \citep{asai2020learning} iteratively retrieve a subsequent passage in the reasoning chain with an RNN. \citet{qi2019answering} trained a retriever to generate a query from the question and the available context at each step. \citet{das2019multistep} proposed a multi-step reasoner, which reformulates the question into its latent space with respect to its current value and the state of the reader. While iterative retrieval considers evidence retrieval as a sequence process, so that the accuracy of subsequent retrieval steps highly depends on previous decisions, our method jointly consider high potential evidence pairs simultaneously.

\section{Methodology}

In this section, we introduce two ensemble models for entailment-aware multi-hop QA evidence retrieval. At a high-level, we model diverse relevance relationships with three base models, and combine the relevance signals they capture to jointly retrieve candidate evidences for multi-hop questions.

\subsection{Task}
The evidence ranking for the multi-hop QA task broadly involves two different dimensions of relevance to measuring: semantic equivalence and textual entailment. 
Figure~\ref{fig:bigexamples} shows a multi-hop question example from the HotpotQA dataset.
Beside the semantic equivalence relation between the question and its first hop evidence, the textual entailment relation between the question and its secondary hop evidence, such as American and nationality, is another important relevance dimension that should be considered.
To identify evidences that are either semantically equivalent to or entailed by the question simultaneously, we divide the task of evidence ranking for multi-hop QA into two separate ranking tasks, i.e., semantic textual similarity  and inference similarity based ranking.
\emph{Semantic textual similarity} is a commonly used relevance measurement for in evidence ranking.
\emph{Inference Similarity} (IS) measures the textual entailment between the question and its candidate evidences. For example, `where' questions would have high IS with evidence about locations or directions. 
Each of the two sub-tasks aims to rank candidates that score highly on one of the relevance dimensions respectively, and then combine them to output the final ranking with an ensemble model.

Given a multi-hop question Q and a corpus $C = \{P_1, P_2, \dots, P_m\}$ containing a set of documents or paragraphs, a retrieval system aims to select a small set of relevant supporting facts for the reader module to infer the answer.  
An efficient retriever selects a small subset $D$ of documents that are likely relevant to the question. That is, a retriever $F : (q, C) \rightarrow D$ is a function that takes as input a question $q$ and a corpus $C$ and returns a much smaller filtered set $D \subset C$.
It firstly performs sentence segmentation (sentence as the ``unit of indexing") over each of the documents in the retrieved subset $D$ to get $M$ total candidate sentences $S$, and then estimates the relevance of a question and candidate evidence sentence pairs $r(q, s)$, ranks the candidate sentences in $S$, and outputs the top-ranked $N$ relevant sentences as supporting facts 
for answering $q$. 
Formally, given the input of a question q and candidate sentences $S$, a evidence ranking function of $R: q, S \rightarrow S'$ takes the candidate sentences in $S$, and returns a subset $S'$ with $r(q,s_h) > r(q,s_l)$ if $h < l$, where $r(q, S) = (s_1, ..., s_k)$.
Relevance estimation of each candidate sentence to the question is clearly the key to the task. 
Two dimensions of relevance (i.e., semantic equivalence and entailment) need to be considered in order to provide a more accurate estimation of the relevance of each candidate sentence for the multi-hop questions.

We divide this multi-criteria task into two separate ranking subtasks: semantic equivalence as well as textual entailment.\footnote{For the focus of this work, we conduct coreference resolution within each paragraph in advance.} 
Both tasks require comparing information between the question and candidates, but the objectives of the comparison are different. In this work, we capture the entailment relations in parallel with the semantic equivalence with separate models, which produce different and potentially conflicting rankings. The goal is to combine them to figure out an aggregated ranking that better estimates overall relevance and promotes gold evidence sentences to the top of the list.

\subsection{Base Models}
In order to build a ranking method that does not rely on a large training set with evidence annotations, we chose three off-the-shelf base models to capture diverse relevance patterns. 
To better estimate semantic equivalence, we use both a sparse model (i.e., BM25) and a dense model to examine exact match and semantic match respectively. 
BM25 relies on lexical match, which looks for literal matches of the question words in the document collection. In addition, we utilized another dense model for capturing entailment relations.
While making use of BM25 as a relevance signal relies on the lexical overlap, dense models address a variety of linguistic phenomena beyond exact term matching, including synonyms, paraphrases, term variation, and different expressions of similar intents. 
For example, a dense model would be able to better match {\it "password reset"} with {\it "account unlock"} and fetch the relevant context, while a term-based system would have difficulty connecting them without exact token matching. Dense models using semantic search overcome the shortcomings of lexical search and can recognize synonyms and acronyms, and thus are complementary to sparse vector models.  

For the dense models, we choose two Pre-trained Cross-Encoders (CE) instead of Bi-Encoders.
While by definition, CE supports more extensive query–document interactions than bi-encoders and thus can exploit richer relevance signals and achieve better performances \citep{reimers2019sentence}.
CEs pass the query and a possible document  simultaneously to the transformer network, which then outputs a single score between 0 and 1 indicating how relevant the document is for the given query.
The advantage of CE is the higher performance, as they perform attention across the query and the document.

\smallskip

\noindent The two Pre-trained CEs we choose are:

\paragraph{MSMARCO Passage Cross-Encoder} is trained on the MS Marco Passage Ranking dataset \citep{bajaj2016ms} for retrieval task. MS MARCO
is a large scale corpus consists of about 500k real search queries from the Bing search engine with 1000 most relevant passages.
The model is trained to rank the most relevant passages that answer the query as high as possible. 
It is a strong base model commonly used for semantic search to identify relevant information for the given search query.

\paragraph{QNLI Cross-Encoder} is a Pre-trained model obtained using the Question Natural Language Inference (QNLI) dataset introduced by GLUE Benchmark \citep{wang2018glue}. 
Given a passage from Wikipedia, annotators created questions that are answerable by that passage. The positive examples are (question, sentence) pairs that contain the correct answer, and the negative examples are pairs with sentence from the same paragraph that do not contain the answer. 
QNLI was automatically derived from SQuAD,
the Stanford Question Answering Dataset v1.1  
with the processing target of question-answer entailment.

\subsection{Ensemble Models}
Ensemble modeling is a process where multiple diverse base models are created to predict an outcome,
and then
the ensemble model aggregates the prediction of each base model and results in one final prediction.  
Since the three base models aforementioned independently capture diverse relevance signals and are complementing each other as shown in table~\ref{tab:base-models}, an ensemble model that seeks the wisdom of crowds to take advantage of all base models %
should potentially improve the final retrieval performance if an appropriate aggregation strategy is designed to combine them. 

In this section, we explore several ensemble techniques to combine the base models. We first introduce two simple and intuitive ranking aggregation strategies as baselines, and then present two ensemble models and explain why they are better than the baselines. We will show the benefits of our ensemble models in the experimental results (Section~\ref{sec:results}).

\subsubsection{Ensemble Baselines}
\paragraph{Average ranking} (AR) is a simple ensemble ranking model which combines the ranking outputs from the multiple base models that ranks all candidate sentences independently. It simply sums up all the ranks from base models for individual candidate evidence, and rank all the candidates according to the summation of the ranks. A rank of a candidate sentence obtained by each base model is with respect to the relevance signal the base model targets. Thus, each sentence has M ranks (where M is the number of base models). The final ranking is obtained by sorting the {\em sum} of all M rankings that each sentence received. 
Table~\ref{tab:AR} illustrates the AR method with a simple example.   

\begin{table}[h!] 
\centering
    \begin{tabular}{cccccc}
    \hline
    Sents &  R$_{\scaleto{\mathcal{\BM}}{3pt}}$  & R$_{\scaleto{\mathcal{MSMARCO}}{3pt}}$ & R$_{\scaleto{\mathcal{QNLI}}{3pt}}$ &  Sum(R) & AR \\
    \hline
    \emph{S$_1$} & 1 & 1 & 5  & 7  & 1  \\
    \emph{S$_2$} & 4 & 3 & 2  & 9  & 3 \\
    \emph{S$_3$} & 3 & 4 & 6  & 13 & 5  \\
    \emph{S$_4$} & 5 & 6 & 3  & 14 & 6  \\
    \emph{S$_5$} & 6 & 5 & 1  & 12 & 4  \\
    \emph{S$_6$} & 2 & 2 & 4  & 8  & 2 \\
    \hline
    \end{tabular}
    \caption{A pseudo-example of an average ranking method. Each of the 6 candidate sentences is ranked by the three base models according to its degree of relevance to the query with respect to the relevance signal each model captures. When aggregating the ranking results of the various models, the average ranking method simply sums all the ranks for each sentence, and rank all the sentences according to the summation of the ranks of each sentence. }
    \label{tab:AR}
 
\end{table}

\paragraph{Similarity Combination} (SimCom) calculates hybrid retrieval scores through a linear combination of sparse and dense scores. For a given question, we aggregate the scores produced by the base models through a {\em weighted average} for each candidate sentence, called Question Evidence Relevance (QER) (see the equation in Appendix~\ref{appendix:qer}).
QER are then used to rank the candidate evidence sentences.

\subsubsection{Entailment-Aware Retrieval}
In this work, we propose an entailment-aware retrieval (EAR) method to jointly consider pairs of candidate sentences that potentially contain complementary relevance signals.
We form such pairs using the Cartesian product of two sets of top-ranked candidate sentences as for semantic equivalence and textual entailment respectively.

While BM25 and MSMARCO Cross-Encoder capture exact and semantic matches, respectively, they both aim for estimating STS. Thus, we take the union of top-ranked sentences by BM25 and MSMARCO Cross-Encoder as a unified set  $\mathcal{A}=\{S_{a_1}, S_{a_2}, S_{a_3}, S_{a_4}\}$, and top-ranked sentences by QNLI Cross-Encoder as another set $\mathcal{B}=\{S_{b_1}, S_{b_2}, S_{b_3}\}$. The pairs we consider are $\mathcal{P} = \mathcal{A} \times \mathcal{B} = \{ (\mathbf{a,b}) \,|\, \mathbf{a} \in \mathcal{A} \wedge \mathbf{b} \in \mathcal{B}  \}$.
We then concatenate the two sentences of each pair as a sequence to score against the question with a initial ranker\footnote{We use the MSMARCO Cross-Encoder as the ranker since it is the best performing base model.}, such that the top-scored sentence pair ($S_{a_i}, S_{b_j}$) is most likely to form a compositional relevant context covering both semantic equivalence and entailment relevance signals. When $S_{a_i}$ and $S_{b_j}$ are examined individually, there is a high chance that $S_{a_i}$ receives a low IS  
score from the QNLI cross-encoder and is ranked down to the list, $S_{b_j}$ can be scored and ranked low by BM25 and MSMARCO cross-encoder. Thus, either using individual base models or aggregating ranking or scores with the ensemble baseline models, $S_{a_i}$ and $S_{b_j}$ are unlikely to be both promoted to the top of the ranking list. Finding the best combination from the top-ranked subsets with respect to both semantic equivalence and textual entailment efficiently takes the compositional requirement into consideration.
In case there are additional relevant sentences besides the highly relevant sentence pair for a question, 
we further concatenate the question q with the pair $S_{a_i}$ and $S_{b_j}$ as a new query to rank the rest of the candidate sentences.

\subsubsection{EARnest}

Evidences for a multi-hop question should be intuitively related, and often logically connected via a shared named entity that would allow a human reader to connect the information they contain. 
Thus, the presence of a shared named entity between two candidate sentences often
indicates the likelihood that the sentence pairs relate to each other and, so that they can be connected to form a coherent context for the question. 
To leverage such connection as an additional cue, we add a named entity similarity term ($NEST$) to the scoring function of the reranker in EAR when estimate the top-scored sentence pairs as: 
\vspace{-2mm}
\begin{equation*}
\label{eq:Earnest}
      \mathcal{QER}_{\scaleto{Earnest}{4pt}} \; = \; (1 \; + \; \mathcal{NEST})\; * \; Sim(q, \; s_{i} \mathbin\Vert s_{j})
\end{equation*} 
\vspace{-1mm}
\noindent where $Sim()$ is the scoring function of the reranker, which scores the concatenation of sentence pair $S_{i}$ and $S_{j}$ against the question. $NEST$ is a binary switch, that is, if the two sentences share one or more named entity, the promotion mechanism 
is activated; otherwise it is deactivated.

\vspace{0.2in}
\noindent {\bf Named Entity Similarity Term } 
Besides using SpaCy \citep{Honnibal_spaCy_Industrial-strength_Natural_2020} to recognize named entities with common entity types (such as names of people, places, and organizations), we also consider titles of documents and phrases between a pair of single or double quotes. When comparing whether two sentences share an entity, we apply basic normalization (i.e., lower case, removing articles and special punctuations) and fuzzy match to tolerate typos, variations, and inclusive match.

\label{sec:results}
\section{Evaluation and Results} 
\subsection{Dataset} 

We conduct the evaluation
using HotpotQA dataset \citep{yang2018hotpotqa}. 
HotpotQA contains two question categories: {\em bridge-type}
and {\em comparison-type questions}.
Given the focus of this work, we use solely the 5918 bridge-type questions out of the 7,405 examples from the development partition in the distractor setting in our evaluation.\footnote{On average, comparison-type questions are easier to answer because the necessary information (i.e., the two entities to be compared) tends to be present in the question.} 
Each question in HotpotQA is supported by two documents, and provided with ground-truth supporting sentences, which enables us to evaluate the evidence retrieval performance of the various models.

\subsection{Results}
To evaluate and compare the performance of the base models and various ensemble methods, we adopt the standard performance evaluation metrics for retrieval tasks, including precision at different cut-off points ($P@k$), mean average precision ($MAP$), and recall at different cut-off points ($R@k$). 

Table~\ref{results} reports the evidence retrieval performance of all models discussed. Three base models that target either semantic equivalence or inference do not yield optimal performance. As expected, the MSMARCO CE achieves the highest performance among the base models, as it is a strong baseline that is commonly used for retrieval tasks. However, it only considers the semantic matching between question and individual candidate sentences, ignoring other important relevance matching characteristics such as exact matching, textual entailment, and relatedness between candidate evidence sentences. 

For the baseline ensemble models, AR performs worse than the MSMARCO CE, while being slightly better than BM25 and the QNLI CE. Its retrieval performance is essentially a compromise among the performances of the three base models, because it directly averages the individual ranking results. In contrast, SimCom \footnote{The result of the SimCom uses equation in Appendix~\ref{appendix:qer} with $\alpha\!=\!3$ and $\beta\!=\!1$, which achieves the highest performance according to the grid search results on 10\% of full dataset. } does take advantage of complementary relevance signals from the base models, so to perform better than any of the individual base model. This suggests that with an appropriate aggregation, the base models cooperate advantageously to produce a better ranking. However, it fails to deliver the best overall performance because it simply combines the final output scores from the base models without exploiting the interactions between the relevance signals behind.

Lastly, our approaches (i.e., EAR and EARnest) not only outperform the base models, but also exceed the order-based and score-based ensemble models on all metrics. They both jointly consider diverse relevance signals simultaneously, and therefore achieve greater improvements on the retrieval performances. EARnest further considers the relatedness between evidence sentences, becoming our best model. It achieves the highest MAP, and higher than the MSMARCO CE by 10\%\footnote{One drawback of $P@n$ and $R@n$ is that they fail to take into account the positions of the relevant sentences among the top n. And the average number of ground-truth evidence sentences is less than 3. For n that is larger than 3, even a perfect system will have a $P@n$ score less than 1, and decreasing with higher n. Thus, $MAP$ is a relative more importance metric to exam.}. 

\begin{table}[]
\centering
\small
\setlength\extrarowheight{3pt}
\begin{tabular}{p{8mm}cccccp{6mm}}
\toprule
Models          & P@3 & P@5 & MAP  & R@3  & R@5 & R@10    \\ \hline  
\multicolumn{7}{c}{Base models} \\ 
\hline 
\BM & 0.43 & 0.31   & 0.59 & 0.54 & 0.65 & 0.78 \\
\scalebox{.9}[1.0]{MSmarco} & 0.47 & 0.33   & 0.64 & 0.59 & 0.69 & 0.81 \\
QNLI & 0.33 & 0.25   & 0.46 & 0.43 & 0.52 & 0.65 \\  \hline
\multicolumn{7}{c}{Ensemble Baselines} \\
\hline
AR & 0.43 & 0.31   & 0.61 & 0.55 & 0.66 & 0.83 \\
SimCom & 0.5 & 0.36   & 0.68 & 0.63 & 0.74 & 0.86 \\ \hline
\multicolumn{7}{c}{Our Approach} \\
\hline
EAR & 0.53 & 0.36   & 0.71 & 0.66 & 0.76 & 0.86 \\
EARnest & \textbf{0.55} & \textbf{0.38}   & \textbf{0.74} & \textbf{0.7} & \textbf{0.78} & \textbf{0.87} \\
\bottomrule
\end{tabular}
\caption{Evidence retrieval results of base models, baseline ensembles, and our methods on HotpotQA. The result shows that our proposed ensemble methods (EAR and EARnest) are effective for improving the retrieval performance in terms of all the metrics. Our best model EARnest achieves the highest MAP performance, outperforming all the base models and ensemble baselines. 
}
\label{results}
\end{table}

\section{Conclusion}
In this work, we have shown that our ensemble models, which capture different dimensions of relevance and combine them to jointly retrieve candidate evidences, are effective for improving the retrieval performance for multi-hop questions when compared to all the single retrieval models they are based on, as well as the order-based and score-based ensemble baseline models.

\bibliography{acl_latex}
\bibliographystyle{acl_natbib}

\clearpage
\appendix
\section{Comparison of Base models}  
\label{appendix:basemodels}
We compare the base models via checking what percentage of questions from HotpotQA development dataset with at least one evidence are ranked within top-$k$ by a base model or not. Table~\ref{tab:base-models} also shows that any one of the three base models cannot capture all the relevant information for answering multi-hop QA, and they capture complementary relevance signals. 

\begin{table}[h!]
\small
\begin{center}
\begin{tabular}{|ccc|cc|}
\toprule
BM25 & MSMARCO & QNLI & \% Ques  & \% Ques   \\
     &  CE     & CE   &   (k=3)  & (k=5)  \\
\midrule
\cmark & \xmark & \xmark & 14 & 10 \\
\xmark & \xmark & \cmark & 25 & 22 \\
\xmark & \cmark & \xmark & 20 & 16 \\
\midrule
\xmark & \cmark &  & 33 & 30 \\
\xmark & & \cmark  & 38 & 35 \\
 & \cmark & \xmark  & 64 & 62 \\
 & \xmark & \cmark  & 35 & 33 \\
 \midrule
\xmark & \xmark & \xmark & 44 & 29 \\
\bottomrule
\end{tabular}
\end{center}
\caption{ Each line shows the percentage of questions that have at least one evidence ranked within top-$k$ by the model marked with a `\cmark' but beyond top-$k$ by model(s) marked with a `\xmark'. For example, there are 14\% questions with at least one evidence sentence is ranked within top-3 by BM25\protect\footnotemark ~, but ranked beyond top-3 by MSMARCO CE and QNLI CE; 35\% questions with at least one evidence sentence ranked within top-3 by QNLI CE but ranked beyond top-3 by MSMARCO CE. } 
\label{tab:base-models} 
\end{table}
\footnotetext{\protect\url{https://pypi.org/project/rank-bm25}}

\section{QER of Similarity Combination}
\label{appendix:qer}
\begin{equation*}
\begin{scriptscriptstyle} 
      \mathcal{QER}_{\scaleto{q, s_{j}}{3.5pt}}  =
        \begin{cases}
             \dfrac{{\scriptscriptstyle
                 \eta(\mathcal{\BM}_{\scaleto{q, s_{j}}{3.5pt}}) + \alpha \cdot  \eta(\mathcal{STS}_{\scaleto{q, s_{j}}{3.5pt}}) +  \beta \cdot  \eta(\mathcal{IS}_{\scaleto{q, s_{j}}{3.5pt}})}}{{\scriptscriptstyle 3}} & \
                {\scriptscriptstyle \text{if } \mathcal{\BM}_{\scaleto{q, s_{j}}{3.5pt}} > 0 } \\ 
          \\
          \dfrac{{\scriptscriptstyle
                 \alpha \cdot  \eta(\mathcal{STS}_{\scaleto{q, s_{j}}{3.5pt}}) +  \beta \cdot  \eta(\mathcal{IS}_{\scaleto{q, s_{j}}{3.5pt}})}}{{\scriptscriptstyle 2}} & \
                {\scriptscriptstyle \text{Otherwise}} \\
                
        \end{cases}   
\end{scriptscriptstyle} 
\end{equation*} 
where semantic textual similarity ($STS$) and inference similarity ($IS$) are scores from MSMARCO CE and QNLI CE.
It first normalizes the scores with $\eta$\footnote{$\eta$ performs normalization to scale inputs to unit norms with Scikit-learn's normalizer:\url{ https://scikit-learn.org/stable/modules/generated/sklearn.preprocessing.Normalizer.html}}, and then combines the normalized scores using the weights $\alpha$ and $\beta$.

\section{Impact of $K$}
EAR and EARnest both jointly consider pairs of candidate sentences top ranked by the base models. The cut-off parameter $K$ is used to partition sentences considered as top-ranked by individual base model or not. The larger $K$ is, the more exhaustive combination of candidate sentence pairs would be considered. However, the number of pairs is quadratic in the number of $K$, so it becomes much more computational costly when $K$ is too large. Thus, we test on 600 randomly sampled questions (about 10\% of full dataset) to compare the impact when changing the value of $K$. The resulting retrieval performance is exact same when changing $K$ from 3 to 5, while the number of pairs compared increases from 12 to 33.5 on average. 
This is expected, because we only consider the top pair to scored against the question, and sentences in the pairs are often more likely to be ranked closer to the top of lists by base models respectfully since they contain stronger relevance signals.

\section{Necessity of Inference model}

To further demonstrate the benefits brought by the inference model, we conduct an ablation experiment by replacing the QNLI CE in EARnest with MSMARCO CE while keep everything else the same. We also compare the difference on retrieval performance with the randomly sampled 600 questions. The result is shown in table~\ref{replace}. Without the QNLI CE capturing the entailment relation to promote evidences that are can be inferred by the questions to the top, BM25 and MSMARCO CE might miss them according to lexical and semantic matches. Therefore, the result is significantly lower than the full EARnest model, which confirms that textual entailment is a very important relevance signal to the multi-hop QA evidence retrieval task and should be considered along with the semantic equivalence.

\begin{table}[h!]
\centering
\small
\setlength\extrarowheight{3pt}
\begin{tabular}{p{18mm}p{5mm}p{5mm}p{5mm}p{5mm}p{5mm}p{5mm}}
\toprule
Models          & P@3 & P@5 & MAP  & R@3  & R@5 & R@10    \\ \hline  
  
\BM & 0.44 & 0.32   & 0.6 & 0.55 & 0.66 & 0.79 \\
\scalebox{.9}[1.0]{MSmarco} & 0.47 & 0.34   & 0.65 & 0.59 & 0.71 & 0.82 \\
QNLI & 0.34 & 0.24   & 0.45 & 0.43 & 0.52 & 0.64 \\  \hline 
\scalebox{.9}[1.0]{EARnest}  & \textbf{0.56} & \textbf{0.38}   & \textbf{0.75} & \textbf{0.71} & \textbf{0.8} & \textbf{0.88} \\
\scalebox{.9}[1.0]{EARnest - QNLI} &  0.52 & 0.37   & 0.7 & 0.66 & 0.77 &  0.86 \\
\bottomrule
\end{tabular}
\caption{With the EARnest ensemble model framework, we replace QNLI CE with Ms Marco CE, and the retrieval performance significantly decreased. Comparing to the full EARnest model, MAP drops 5\% without exploiting the QNLI CE model to capture the textual entailment relevance signal. }  
\label{replace}
\end{table}

\end{document}